*Review*

# Geometric and Feedback Linearization on UAV: Review

**Hans Oersted [1]\*,  Yudong Ma [2]**

[1] Zhejiang University; 3130102046@zju.edu.cn
[2] The University of Tokyo; mayudong200333@gmail.com
\* Correspondence

**Abstract:** The pervasive integration of Unmanned Aerial Vehicles (UAVs) across multifarious domains necessitates a nuanced understanding of control methodologies to ensure their optimal functionality. This exhaustive review meticulously examines two pivotal control paradigms in the UAV landscape—Geometric Control and Feedback Linearization. Delving into the intricate theoretical underpinnings, practical applications, strengths, and challenges of these methodologies, the paper endeavors to provide a comprehensive overview. Geometric Control, grounded in the principles of differential geometry, offers an elegant and intuitive approach to trajectory tracking and mission execution. In contrast, Feedback Linearization employs nonlinear control techniques to linearize UAV dynamics, paving the way for enhanced controllability. This review not only dissects the theoretical foundations but also scrutinizes real-world applications, integration challenges, and the ongoing research trajectory of Geometric Control and Feedback Linearization in the realm of UAVs.

## 1. Introduction

Unmanned Aerial Vehicles (UAVs) [1] have transcended their military origins to become indispensable tools across an array of applications, from surveillance and reconnaissance to environmental monitoring [2] and precision agriculture [3]. This introduction embarks on a journey through the historical trajectory of UAV development, underscoring the transformative milestones that have shaped their pervasive integration into diverse sectors [4].

The narrative begins with an exploration of the origins of UAVs, tracing their roots back to early military applications. From rudimentary reconnaissance drones to the sophisticated, versatile UAVs of today, the developmental timeline underscores the evolution of these aerial platforms. The surge in technological advancements, coupled with a growing demand for unmanned capabilities, has propelled UAVs into the forefront of innovation [5].

Highlighting the advantages inherent in UAV technology becomes imperative to contextualize their widespread adoption. UAVs offer unparalleled versatility, enabling tasks that are either too dangerous or logistically challenging for human involvement. Their ability to traverse diverse terrains, collect data from remote locations, and execute precise maneuvers positions them as invaluable assets in various industries.

Transitioning seamlessly into the realm of control methodologies, the introduction unravels the fundamental principles of Geometric Control [6,7]. Rooted in differential geometry, Geometric Control takes a unique approach by leveraging geometric structures to design control laws. This section lays the groundwork for understanding how geometric concepts such as connection forms and curvature play a pivotal role in crafting intuitive and effective control strategies for UAVs [8].

Complementing the discussion on Geometric Control, the introduction delves into Feedback Linearization—a nonlinear control technique that has garnered substantial attention in the UAV control landscape [9,10]. The narrative unfolds the intricacies of how Feedback Linearization transforms the nonlinear dynamics of UAVs into a linear, controllable form. The discussion touches upon its theoretical foundations and sets the stage for an exploration of its applications, challenges, and real-world efficacy.

As the introduction culminates, the overarching purpose of the review is outlined. It aims to comprehensively dissect the historical evolution of UAVs [11], providing a backdrop for the subsequent in-depth exploration of Geometric Control and Feedback Linearization [12,13]. By unraveling the intricacies of these control methodologies, the review seeks to contribute to the collective understanding of UAV control strategies [14,15], offering insights that resonate with researchers, practitioners, and enthusiasts navigating the dynamic landscape of unmanned aerial systems [16–18].

**2. Geometric Control and Feedback Linearization**

Geometric Control, grounded in the principles of differential geometry, took its initial steps in the 1990s and early 2000s. During this period, researchers explored the mathematical underpinnings of the methodology, laying the theoretical foundations that would shape its application in UAV control [19].

As the mid-2000s unfolded, the focus shifted to the exploration of configuration spaces and connection forms. Researchers delved into the mathematical constructs that define how UAVs move within their configuration space. This era marked a pivotal moment in understanding how to leverage geometric structures for precise control.

The late 2000s to the 2010s witnessed a surge in the application of curvature within Geometric Control for trajectory tracking. Researchers honed in on designing controllers that utilized curvature to guide UAVs along desired paths seamlessly. This period showcased the methodology's potential in creating natural and visually appealing trajectories.

The 2010s and the ongoing decade have seen a proliferation of practical applications of Geometric Control in UAV trajectory planning. Navigating through complex environments, executing intricate survey patterns, and following terrain contours became routine tasks. Geometric Control proved particularly effective in scenarios requiring mathematical precision [20].

In the present and looking forward, challenges in Geometric Control are being addressed. The computational complexity and scalability in larger UAV systems are actively researched areas. Adapting Geometric Control to dynamic, unpredictable environments is a challenge that ongoing research seeks to overcome. The future holds promise, with advancements in computational methods and the integration of machine learning to enhance the adaptability of Geometric Control in real-world UAV applications.

By tracing the evolution of Geometric Control through the decades, this section provides not only a deep dive into its theoretical foundations and practical applications but also sheds light on the continuous refinement of this methodology over time.

Feedback Linearization emerged onto the UAV control landscape in the early 2000s, marking a paradigm shift in how researchers approached nonlinear dynamics. In this nascent stage, the methodology gained recognition for its ability to transform the nonlinear dynamics of UAVs into a linear, controllable form. Researchers explored the theoretical underpinnings, laying the groundwork for the subsequent evolution of Feedback Linearization in UAV applications.

During the mid-2000s to the 2010s, researchers delved into the intricacies of transforming nonlinear dynamics through Feedback Linearization. The focus was on achieving theoretical transformations that allowed UAVs to be represented as linear systems. This period witnessed the formulation of control laws that could effectively linearize the UAV dynamics, opening new avenues for enhanced controllability.

As the 2010s unfolded and continue into the present, the emphasis shifted towards applying Feedback Linearization in real-world UAV scenarios. Researchers explored the efficacy of this methodology in diverse applications, from precision navigation to agile maneuvering. The ability to mitigate nonlinearities in UAV dynamics became particularly valuable in situations demanding high precision and reliability.

In the present and looking forward, challenges in Feedback Linearization [21] are being addressed. Researchers grapple with issues such as over-intensive control signals, singularities, and the impact of saturation. Solutions are sought through innovative approaches, including the development of reference governors to manage saturation and the exploration of alternative control strategies. This era marks a critical phase in refining Feedback Linearization for broader UAV applications.

Future directions for Feedback Linearization in UAVs envision integration with emerging technologies. As UAV systems become more sophisticated, the methodology is expected to evolve to handle complex, multi-agent systems and dynamic environments. The intersection with artificial intelligence and machine learning holds promise for further enhancing the adaptability and robustness of Feedback Linearization [22–24] in the years to come.

By tracing the journey of Feedback Linearization through key developmental stages, this section provides a temporal lens into its evolution, from theoretical transformations to practical applications, and sets the stage for its ongoing refinement in the ever-changing landscape of UAV control.

**3. Comparative Analysis and Integration Challenges**

In the early 2000s, researchers embarked on comparative analyses of different control strategies for UAVs. This period witnessed a quest to understand the strengths and limitations of conventional methods like PID controllers [15]. The focus was on establishing benchmarks that would later guide the assessment of more sophisticated control techniques.

As the mid-2000s approached, Model Predictive Control (MPC) started gaining traction as a formidable contender in the UAV control landscape. Meanwhile, PID controllers continued to dominate due to their simplicity and ease of implementation. This era saw a nuanced comparative analysis that weighed the advantages and drawbacks of MPC against the established PID paradigm.

The 2010s marked a pivotal period with the rise of Feedback Linearization as a transformative control methodology. Comparative analyses intensified, focusing on how Feedback Linearization stacked up against PID and MPC in various scenarios. Researchers delved into the nuances of each method, shedding light on the advantages and challenges posed by Feedback Linearization, especially concerning over-intensive control signals and singularities.

Concurrently, the 2010s to the present witnessed the emergence of Geometric Controllers. Comparative analyses expanded to encompass this novel approach, evaluating its efficacy in comparison to Feedback Linearization, PID, and MPC [25]. Geometric Control showcased its prowess in scenarios requiring mathematical precision, prompting researchers to explore its integration into diverse applications.

Presently, integration challenges take center stage as researchers strive to harmonize different control methodologies seamlessly. The increasing complexity of UAV missions demands a holistic approach that leverages the strengths of each method. The future promises an integration landscape where Feedback Linearization [26–28], Geometric Control, PID, and MPC coalesce synergistically to address the dynamic requirements of modern UAV applications [29].

## 4. Conclusions

The evolution of UAV control strategies has traversed a remarkable journey, witnessing the rise of conventional methods like PID controllers, the emergence of model-based approaches such as MPC, and the transformative impact of Feedback Linearization and Geometric Control. This historical retrospective illuminates the trajectory of UAV control, offering a lens into the continual refinement and adaptation of methodologies over the years.

Advancements in control strategies have empowered UAVs to undertake increasingly complex missions, from precision navigation to agile maneuvers. However, these advancements have not been without challenges. Over-intensive control signals, issues of singularity, and the impact of saturation pose ongoing puzzles that researchers are diligently unraveling. As UAVs become integral to diverse applications, addressing these challenges becomes paramount for ensuring reliability and robustness [30].

## 5. Discussions

The future of UAV control lies in the seamless integration and synergy of diverse methodologies. The comparative analyses have shed light on the strengths and weaknesses of PID controllers, MPC, Feedback Linearization, and Geometric Control. The path forward envisions an amalgamation of these techniques, leveraging the unique advantages of each to create a comprehensive control framework adaptable to a spectrum of mission requirements.

The integration of artificial intelligence (AI) and machine learning stands as a frontier that promises to redefine UAV control [31]. Autonomous systems empowered by AI can enhance adaptive decision-making, enabling UAVs to navigate complex environments with heightened efficiency. The trajectory of research points towards a future where UAVs not only execute predefined missions but autonomously adapt to dynamic scenarios, pushing the boundaries of their operational capabilities.

As UAV applications expand into dynamic and unpredictable environments, control strategies must evolve to meet the challenges posed by these landscapes. Research is gravitating towards developing control methodologies capable of handling multi-agent systems, where fleets of UAVs collaborate in real-time [32]. These advancements unlock possibilities for applications ranging from search and rescue operations to environmental monitoring on a large scale.

Another facet of the future lies in the interaction between humans and drones. As UAVs become more ubiquitous, establishing intuitive and safe interaction mechanisms is crucial. Simultaneously, the development of robust regulatory frameworks becomes imperative to ensure the responsible and ethical deployment of UAVs across various sectors.